# Vision-based Driver Assistance Systems: Survey, Taxonomy and Advances

Jonathan Horgan, Ciarán Hughes, John McDonald, Senthil Yogamani
Computer Vision Department, Automated Parking
Valeo Vision Systems, Ireland
senthil.yogamani@valeo.com

***Abstract*—Vision-based driver assistance systems is one of the rapidly growing research areas of ITS, due to various factors such as the increased level of safety requirements in automotive, computational power in embedded systems, and desire to get closer to autonomous driving. It is a cross disciplinary area encompassing specialised fields like computer vision, machine learning, robotic navigation, embedded systems, automotive electronics and safety critical software. In this paper, we survey the list of vision based advanced driver assistance systems with a consistent terminology and propose a taxonomy. We also propose an abstract model in an attempt to formalize a top-down view of application development to scale towards autonomous driving system.**

***Keywords*—Automotive Vision; Embedded Vision; Autonomous Driving; ADAS; Machine Learning; Computer Vision**

## I. Introduction

Most automotive manufacturers employ some form of camera system, at least on their high end models, whether it is a simple rear-view camera for back-over protection, front camera for lane departure warning, front collision warning, etc., or stereo cameras for more complete depth estimation of the environment ahead of the vehicle. Naturally, the more advanced applications require higher-end processing units, and as such incur a higher production cost to the manufacturers. However, as safety legislation for automobiles advances, camera systems will increasingly feature in lower-end vehicles, and soon will be a standard fit for many vehicle variants.

Fig. 1 gives the high-level coverage of several current vision-based applications, from viewing applications such as Surround View (SV), to driver assistance system such as Cross Traffic Alert (CTA), Park Assist (PA), and Collision Warning (FCW and RCW). Several safety critical vision systems are also in production (or soon will be), such as Adaptive Cruise Control (ACC) and electronic mirror replacement (CMS). What is immediately obvious is that camera systems are among the most versatile sensor systems available for the automotive market, offering the broadest range of applications of any sensor system.

The aim of this paper is to give a high-level survey of the technology, vision application development and current & future technologies. This will include currently available vision-based products, along with next generation and emerging applications in automotive vision systems [1][2][3]. Finally, we will discuss the trend towards the use of vision systems in autonomous vehicles, which is a particularly pertinent topic, as in many jurisdictions legislators are starting to implement laws that allow autonomous vehicles on public roads. In so doing, they are acknowledging the fact that autonomous vehicles will revolutionise the automotive industry worldwide, by improving road safety, traffic efficiency and the general driving experience. We will discuss the need to model and formulate the problem in a generic framework, which is robust and scalable in order to support different applications. We will also propose a generic scalable software model for vision-based ADAS development.

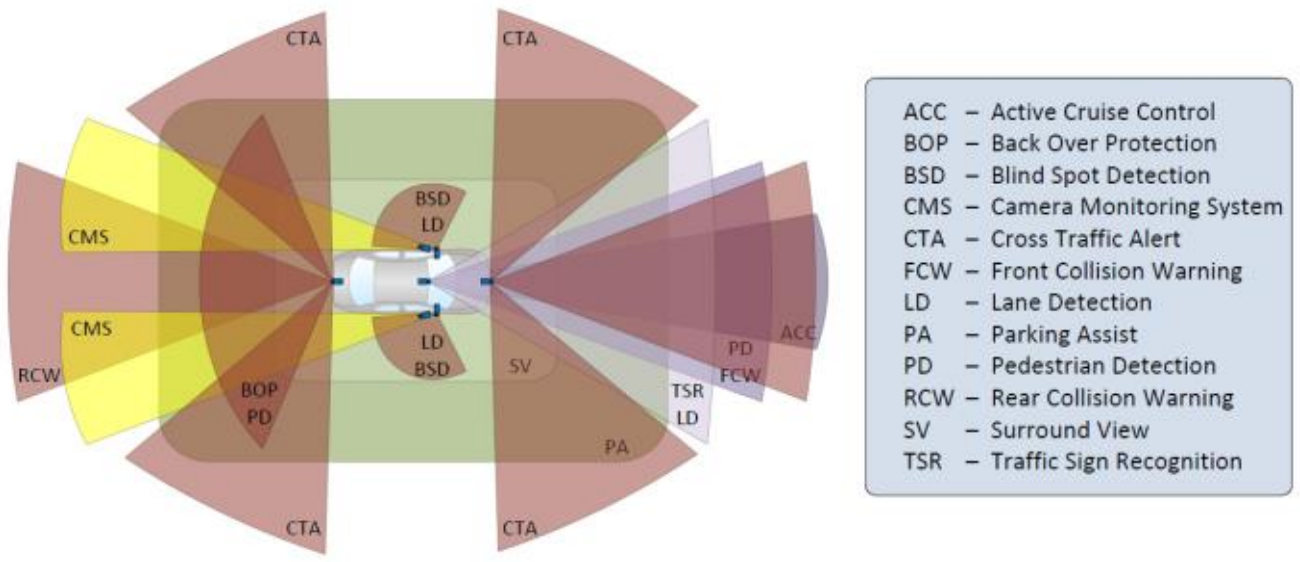

Fig. 1. Camera system interconnect and application field of view.

## II. Taxonomy

In this section, we provide different ways of classifying vision systems and their usage in delivering ADAS functionality. In the current literature, there is no consistent terminology and classification for the applications discussed in this paper. Here we propose a consistent and precise terminology for vision based ADAS.

**Display (A_DI)/Alert (A_AL)/Semi-Automated (A_SA) / Automated (A_AD)/Autonomous (A_AS):** This classification describes the level of interaction between the vision system and the user/system itself. This breakdown is similar to the levels of automation proposed by NHTSA [4]. If a function reaches the autonomous level, it can also be employed as a function at a lower level of automation.

When cameras were first fitted to production vehicles almost 25 years ago, they were simply used as an aid to improve driver visibility in difficult scenarios, such as reversing or night vision. Different view ports can be rendered using individual camera inputs and overlays such as the driving tube, can further aid in positioning of the vehicle. Advances in the number of cameras, and the hardware allowed for the stitching of four camera inputs, provide the user with an almost seamless 2D top view. This provides driver aid in many more driving situations, such as parking and low speed manoeuvring close to objects. The most recent advancement in viewing is the creation of a 3D bowl view around the vehicle, which allows the user to change the view point of a virtual camera.

While viewing is the primary function of surround view cameras (front, rear, left and right mirror), front view cameras as they are known (typically mounted behind the rear-view mirror), are targeted solely to provide driver aids and typically have no direct visual output to the user. Cameras have become essential sensors in standard alert type functions such as traffic sign recognition (TSR) [5][6], cross traffic alert (CTA) [7], lane departure warning (LDW) [8] and object detection (OD) [7]. These functions provide the driver with valuable information about the driving situation, but it is up to the driver whether they act on this information or not.

Semi-automated systems provide functions in which the driver is still fully responsible for overall control of the vehicle, but one function (braking/steering/light switching) is automated. The driver can assume control of the automated function at anytime. Examples of such functions include adaptive cruise control (ACC) [9], automatic emergency braking (AEB) [10], lane keeping (LK) [11], high beam assist (HBA) [12].

Automated systems allow the driver to hand over control of more than one safety critical function for vehicle navigation (accelerating/braking/steering) at once, under certain conditions and rely on the vehicle to safely navigate the environment itself. Control of the vehicle is returned to the driver on request, or completion of the task, or when the system recognises a situation that cannot be automated safely. The driver is still expected to be attentive and ready to take over from the system at reasonably short notice. Examples of such systems include, automated parking (AP), highway driving assist (HDA), and traffic jam assist (TJA).

Autonomous driving is the end goal for the industry. Such automation would allow the user to simply input a destination and then, not be expected to take over control of the vehicle for the duration of the journey. These systems will consist of many of the functions mentioned above, functioning as part of an overall autonomous driving system. The fully autonomous system will also be able to operate in a driver-less fashion. This category can be further subdivided, as there will be levels of autonomous driving that will be introduced in steps over the coming years.

The first systems that will be available on the market will be partially autonomous. These systems will take over full vehicle control for large portions of the journey and return control to the driver (with an appropriate amount of notice), in situations in which autonomous driving is more difficult. Eventually, these difficult situations will be handled and fully autonomous systems will come to the market. It should be noted that certain levels of infrastructure may be expected in order to achieve this.

**Interior (V_I)/Exterior (V_E):** A broad categorisation of automotive camera systems is whether they are interior or exterior cameras. This categorisation refers to where the field of view of the camera is covering, rather than the physical location of the camera on the vehicle. Typical automotive cameras are exterior facing, providing information to the user (or vehicle systems) about the external environment in which vehicle is navigating. This system can provide applications such as CTA and TSR, among others. In more recent years, there is a trend towards the addition of interior monitoring cameras. The purpose of these systems is typically for monitoring the occupants of the car (the driver in particular), providing functions such as driver monitoring (DM), back-seat child monitoring [13] and gesture recognition [14].

**Single/Surround View (SV):** Traditional rear/ front/ night vision camera solutions are handled as single sources, having no interaction with other onboard camera systems. Front single cameras are typically narrow field of view (FOV) cameras (<60 degrees), whereas reversing cameras can have wider FOVs. Surround view systems consist of four fisheye cameras (>160 degrees FOV) (front view (mounted in front grill), left and right mirror view (mounted under wing mirrors) and rear view (typically mounted under boot lip)), connected to a single ECU [15]. These cameras are controlled from an acquisition, calibration and image quality perspective to allow generation of views such as top-view and 3D bowl-view, while also allowing for image processing functions. Such functions can take advantage of either single or multiple views, such as lane marking detection and object detection (OD).

**Mono/Stereo(ST):** The vast majority of automotive camera systems are monocular. In more recent years, some automotive suppliers have turned to stereo systems. A stereo camera consists of two or more lenses with a separate image sensor for each lens. These, effectively separate cameras, are paired together with a known baseline between the cameras, and an overlapping FOV. For front camera applications this allows for more accurate depth estimation in the scene, increased robustness and provides a level of redundancy. Stereo cameras can enhance functions such AEB and ACC. These systems are more expensive than monocular ones due to both, the extra camera and processing hardware required, while also having greater package size.

**Active (C_A)/Passive (C_P):** Typical automotive cameras are passive systems. They use the ambient light in the environment and operate within the visible electromagnetic spectrum. Therefore, system performance generally degrades in low light conditions. There are active automotive cameras in the market, such as active night vision (NV) [16] and time of flight (TOF) [17] cameras, that require an artificial infrared light source to illuminate the scene. The wavelength of the light source in active systems makes it invisible to the human eye. These cameras are typically focused on providing ADAS functions to the user, rather than being used for display purposes, though some manufacturers provide option for display.

**Low Speed (S_L)/High Speed (S_H):** Cameras used for purely viewing/low speed manoeuvring (<20Kph) purposes, tend to operate at a frame rate between 25 and 30 frames per second (fps) (often guided by the requirements of the head unit display). The frame rates of these cameras can be further reduced (e.g. 15fps) in low light conditions, to improve image quality by increasing exposure time. Cameras used for high speed applications such as ACC or AEB and cameras used for viewing at high speeds, such as mirror replacement (CMS) cameras can operate up to 60fps or higher. Vision applications running on all of these cameras, can process either, all captured images or a have a lower frame rate multiple of the camera capture rate. The selection of algorithm processing frame rate depends on many factors, including the vehicle speed during operation of the function, requirements for inter-frame motion, the computational load of the algorithm and the update rate required from the function itself.

**Standalone/Fusion (FUS)**: Standalone cameras are systems that provide the relevant output to the user without any fusion of data with other sensor systems. This is the case for most display cameras, as fusion is not necessary, but it is also the case with most currently available front/rear and surround view solutions. There are fusion systems available on the market at present that fuse both Lidar and Radar sensor data, along with visual information to improve system availability, accuracy and robustness. Sensor fusion remains a relatively young area of research in the automotive sector; however it is rapidly growing due to its recognised importance in achieving the goals of automated driving. Multi-sensor data fusion from cameras, ultrasonics, lidar and radar sensors will allow next generation systems to achieve the availability, accuracy and robustness required to accomplish the aim of automated driving.

## III. Vision Algorithm Building Blocks

In this section, we discuss the basic building blocks of automotive vision ADAS applications for standard monocular and stereo viewing cameras. This section does not cover night vision and time of flight cameras, but many steps are still applicable. These topics are covered in detail by computer vision textbooks and open-source software libraries [18][19] and here we summarize it according to the needs of typical vision applications. A combination of a subset of the following image processing techniques, along with high-level interpretation logic form the basis for most image processing based ADAS applications.

**Acquisition:** Firstly, an image is captured and presented to the image processing algorithm for processing at a particular predefined frame rate. The function can process purely greyscale images, or may also require colour information depending on the algorithm implementation.

**Controller:** Prior to processing, it is usual that some checks are performed on the image to ensure that both the image and the function itself are in a state capable of providing useful results. These checks could include some of the following (if the relevant information is available on the system): calibration accuracy checks, image soiling check and image quality checks, such as noise and light level.

**Pre-processing:** The purpose of this step is to improve the acquired image, in order to increase the effectiveness of subsequent processing. Pre-processing can consist of many different operations including: region of interest extraction, image down sampling, image pyramid generation (for handling scale), image smoothing (such as Gaussian for reducing image noise) or edge detection (such as Sobel for highlighting image gradients).

**Segmentation:** Segmentation is the process of partitioning an image into multiple areas of similar characteristics. The purpose is to simplify the representation of an image by locating object boundaries (such as road edge or sky), which can result in reduced processing time and improved performance of the subsequent processing steps [20]. Watershed segmentation is an example of a commonly used approach [21].

**Motion Estimation:** This is the basis for many interesting ADAS applications, and is often the most computationally expensive part of the algorithm. Motion estimation involves the tracking or matching of image pixels from one image to the next image in time. The motion of pixels on the image is as a result of two motions; 1) the corresponding 3D point is moving in the scene and/or 2) the camera itself is moving relative to the scene. Tracking can be subdivided into two main methods, namely sparse and dense. Sparse tracking consists of first detecting salient features on the image (such as corners, e.g. Harris corner detector [22]), and then tracking them using optical flow techniques (e.g. Lucas Kanade tracker [23]). Dense optical flow involves tracking the location of each pixel in the image (e.g. Horn–Schunck [24]). Both methods have their own benefits: sparse tending to provide a higher percentage of matched points, with a lower processing cost, while dense optical flow provides greater image coverage and more statistical information to handle outliers. Matching methods perform detection and description of features (salient pixels), which are then matched to features detected in the subsequent image. Examples of such feature detectors include A-KAZE, SIFT, SURF and ORB. These methods are more robust than optical flow to image noise, lighting changes and projective transformations; however, they are sparser in the image and more significantly, they are more computationally expensive to calculate.

**Depth Estimation:** Depth estimation refers to the set of algorithms aimed at obtaining a representation of the spatial structure of the environment within the sensor's FOV. While this is the primary focus of many active sensor systems, such as TOF cameras, lidar and radar, this remains a complex topic for passive sensors. There are two main types of depth perception techniques for cameras: namely stereo and monocular [25].

The primary advantage of stereo cameras over monocular systems is improved ability to sense depth. It works by solving the correspondence problem for each pixel, allowing for mapping of pixel locations from the left camera image to the right camera image. The map showing these distances between pixels is called a disparity map, and these distances are proportional to the physical distance of the corresponding world point from the camera. Using the known camera calibrations and baseline, the rays forming the pixel pairs between both cameras can be projected and triangulated to solve for a 3D position in the world for each pixel.

Monocular systems are also able to sense depth, but motion of the camera is required to create the baseline for triangulation. This method of scene reconstruction is referred to as structure from motion (SFM). As described in the motion estimation section above, pixels in the image are tracked or matched from one frame to the next using either sparse or dense techniques. The known motion of the camera between the processed frames as well as the camera calibration, are used to project and triangulate the world positions of the point correspondences. Bundle adjustment is a commonly used approach to simultaneously refine the 3D positions estimated in the scene and the relative motion of the camera, according to an optimality criterion, involving the corresponding image projections of all points. Example of this is shown in Fig. 2.

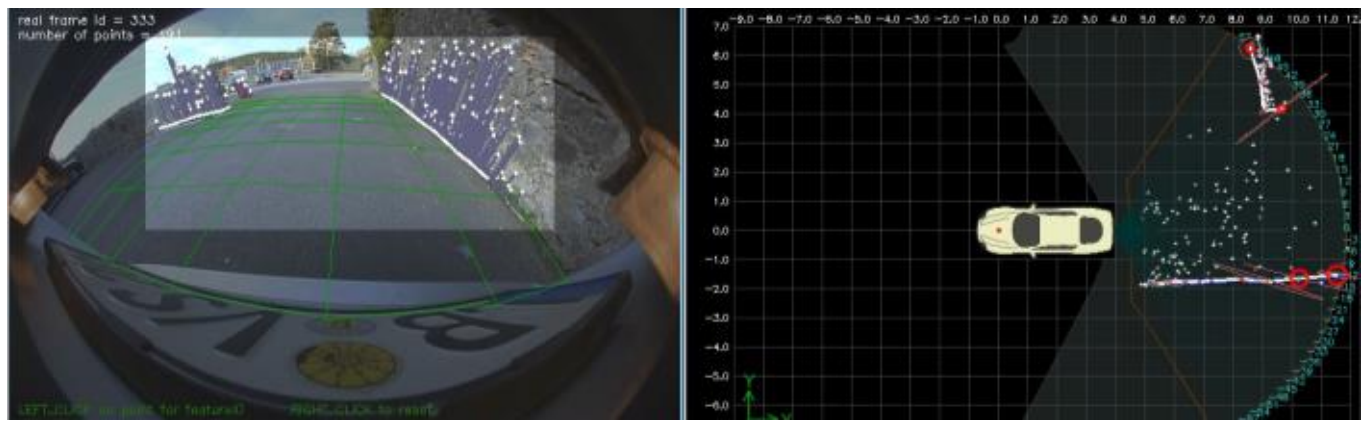

Fig. 2. Depth estimation from monocular camera

Stereo approaches are more accurate, as they have a known fixed baseline and use smaller FOV cameras which allows for better spatial resolution in the processed area. They also require less intensive processing as the correspondence problem is simpler to solve; however, the 3D reconstruction is limited to the overlapping area of the two cameras and the system cost of a stereo camera is higher. SFM techniques usually employ sparse processing to reduce computational complexity. For mirror cameras, the SFM problem is similar to stereo because of the lateral motion creating a virtual baseline (motion stereo). However, this is complicated for front cameras due to the longitudinal motion of the camera relative to the scene, resulting in poor triangulation near the focus of expansion in the image.

**State Tracking:** State tracking can be employed in many different places throughout the image processing pipeline to improve performance. State tracking can be used in order to provide a level of robustness against missed/dropped detections and noise, handle erroneous calculations, predict the state at next time interval, and also, to temporally smooth the output of different parts of the processing chain. Common examples of such state trackers include the Kalman filter [26] and Particle filter [27].

**Object detection/Classification:** There are two main types of classification: namely supervised and unsupervised. Supervised classification is based on the knowledge that a human can select sample thumbnails in multiple images that are representative of a specific class of object. With the use of feature extraction methods such as histogram of oriented gradients (HOG), local binary patterns (LBP), and wavelets applied to the human classified sample images, a predictor model is built using machine learning in order to classify objects. Many vision based ADAS functions use machine learning approaches for classification in applications such as pedestrian detection (PD) [28], face detection, vehicle detection and traffic sign recognition (TSR). The quality of the final algorithm is highly dependent on the amount and quality of the sample data used for learning the classifiers. Typical classifiers include SVMs, random forest and convolutional neural networks.

Unsupervised classification is the problem of trying to find structure in data without having prior knowledge of the data to be classified. Popular approaches to unsupervised classification in image processing include K-mean clustering and Iterative Self-Organizing Data Analysis Technique (ISODATA). Unsupervised classification is typically used as part of other applications, such as the clustering of moving objects.

## IV. Survey of Applications

### A. *Established Applications*

Vision based ADAS applications first started appearing in mass production in early 2000s with the release of systems such as lane departure warning (LDW) [8]. Since then, there has been rapid development in the area of vision based ADAS. This is due to the vast improvements in processing and imaging hardware, and the drive in the automotive industry to add more ADAS features in order to enhance safety and improve brand awareness in the market. As cameras are rapidly being accepted as standard equipment for improved driver visibility (surround view systems), it is logical that these sensors are employed for ADAS applications in parallel.

Some established vision based ADAS applications are listed below, along with their taxonomy and the typical vision building blocks required as detailed in the previous two sections. This is not a complete list of all vision applications and most solutions can be developed in many different ways. The taxonomy detailed in the Table 1 assumes that the applications can use passive, monocular, standalone camera systems unless otherwise stated. Vehicle manufacturers use different names for ADAS systems that are providing the same functionality; this can cause confusion as many functions have multiple names. When including surround view in the taxonomy, it means the application could either run on a single surround view camera or use all four.

TABLE I. TAXONOMY OF VISION BASED ADAS APPLICATIONS

| Application | Typical Vision Building Blocks | Taxonomy |
|---|---|---|
| Surround View System (SVS) / 3D view | No significant image processing required. May be employed to improve image stitching accuracy. Allows for overlaying | A_DI, V_E, SV, C_P, S_L, FUS |
| Adaptive Cruise Control (ACC) | Supervised Classifier (Vehicle rear), Segmentation, State Tracking | A_SA, V_E, SV, ST, S_H, FUS |
| Lane Departure Warning (LDW) / Lane Keeping System (LKS) | Lane detection based on edges, color and curve, Segmentation, State Tracking | A_SA, V_E, SV, S_H, FUS |
| Adaptive Headlights (AH) / High Beam Assist (HBA) | Supervised Classifier (Vehicles rear and front lights), State Tracking, Motion Estimation | A_AD, V_E, S_H |
| Night Vision Systems (NVS) / Night Vision | Passive or Active Infrared, Segmentation | A_DI, V_E, C_A, S_L |
| Cross Traffic Alert (CTA) | Motion Estimation, Segmentation, Unsupervised Classifier, State Tracking | A_SA, V_E, SV (requires fisheye), S_L, FUS |
| Forward Collision Warning (FCW) / | Supervised Classifier (Vehicle rear), State Tracking | A_SA, V_E, SV, ST, S_H, FUS |
| Traffic Sign Recognition (TSR) | Color filtering, Supervised Classifier (Signs from different countries) | A_AL, V_E, SV, S_H |
| Object Detection (OD) / Pedestrian Detection (PD) / Back Over Protection | Motion Estimation, SFM, Supervised Classifier (Pedestrians and vehicles), Segmentation | A_AL, V_E, SV, ST, S_L, FUS |
| Driver Monitoring System (DMS) / Attention Assist (AA) | TOF or Passive camera, Background Subtraction,Supervised Classifier (Face detection), Eye Tracking | A_AL, V_I, C_A |
| Back-seat Children monitoring | Background Subtraction, Supervised Classifier (Face detection) | A_AL, V_I, |
| Traffic Light Recognition (TLR) / Traffic Light | Color filtering, Supervised Classifier (Traffic lights from different countries) | A_AL, V_E, SV, S_H |

## *B. Emerging Applications*

While the established applications mentioned above are still being further developed to improve robustness, accuracy and availability, there are a number of new ADAS applications in development. These new driver aids typically require a combination of the established applications, plus some newly developed vision functions as well as fusion with other sensor technologies. Here we discuss just some of the many ADAS applications that will be coming to the market over the next few years.

**Automated Parking (AP):** Automated parking systems have been on the mass market for some time. This started with automating parallel parking and then moved to include perpendicular parking systems. The systems are now evolving beyond the semi-automated types in which only steering is controlled, to becoming fully automated. These fully automated systems, which are now becoming available, allow the driver to exit the vehicle and initiate the parking manoeuvre remotely through a key fob or smart phone. In this case, the driver remains responsible for monitoring the vehicles surroundings (at all times), and the parking manoeuvre is controlled through a dead man switch on the key fob or smart phone. They are appropriate in scenarios where the parking space has already been located and measured or in controlled environments, (such as garage parking), where the vehicle can be safely allowed to explore the environment in front with limited distance and steering angle.

The next step for parking systems is to make them truly autonomous, which will allow a driver to leave a car to locate and park in an unmapped environment without any driver input. In addition to this, the vehicle should be able to exit the parking slot and return to the driver safely. Cameras can play a very important role in the future of automated parking systems, providing important information about the vehicles surroundings. This includes information like object and freespace data, parking slot marking detection, pedestrian detection for fusion with other sensor technologies.

**Traffic Jam Assist (TJA)/ Highway Driving Assist (HDA):** TJA and HDA are automated systems for handling both steering (lateral control) and acceleration (longitudinal control) in driving situations. Both functions can also be known by different names such as traffic jam pilot and highway pilot, but the basic function remains the same. The difference between the two functions is the speed of the vehicle and the driving situation they are intended for; TJA is intended for low speed manoeuvring in congested traffic situations, while HDA is intended for use on high speed driving. As these are automated systems, the driver is expected to monitor the driving situation and be prepared to take full control of the vehicle at short notice. Some systems monitor driver awareness as a requisite for system functionality. If it recognises that the driver is not paying attention, then the system will warn the driver and hand back full control. An example of this monitoring is ensuring that the driver's hands are on the steering wheel.

TJA and HDA combine the functionality of following the leading vehicle at a safe distance and lane keep, all while staying within the speed limit. TJA is also responsible for stopping and starting of the vehicle. Initially, both are targeted for operation in limited access highways. They both operate in a similar way, using a combination of the functions mentioned above including ACC, LK and TSR in order to control the vehicle. Sensor fusion will be used to achieve these functions to ensure robustness, availability and sensor redundancy. They can take advantage of either monocular or stereo vision systems. These will be the first such systems on the market in which the driver will hand over control to the ADAS system for driving situations, while still monitoring its operation. As such, these functions will act as an important step in gauging public perception of automated driving, while also providing valuable usage statistics.

## V. Towards autonomous driving - Challenges and future directions

### A. Challenges and other cues

Object detection algorithms in computer vision are solutions of inverse problems as they recover semantics from 2D image projections. Object modelling is difficult because of the variations due to lighting, pose, colour and texture. Due to these difficulties, machine learning has become dominant. The recent trend of deep learning demonstrates how generic models with no prior knowledge of the object can outperform others especially when the dataset and number of objects are large. In particular, an automotive vision system has its own set of unique challenges. There are multiple cameras typically one on each side of the car working collaboratively. In V2V (Vehicle to Vehicle) systems, the cameras in different cars work collaboratively forming a multi-camera network. The cameras mounted on the car are moving in a dynamic environment with other objects motion with or without correlation. The infrastructure around is not known a priori. Wide-angle fish-eye lenses are used for covering a larger area which causes lens distortion which breaks typical geometric models based on pin-hole cameras. On the other hand, there is well defined structure where-in the focus of attention is on the area in front of the car (referred to as free space) and the objects interacting near the vicinity of the cameras (leading vehicle). The motion of the cameras can be estimated via inertial sensors of the car (steering angle and wheel RPMs).

**Location priori:** GPS in cars provides its accurate location. Location can be used a strong priori providing information about the static infrastructure in the scene. Google's Street View is one such service which can provide prior recordings of the scene's infrastructure. Downloading this data continuously could increase the data-traffic needs and also the power consumption. Hence there is a trend of HD maps which stream only critical information like road boundaries, lanes, etc rather than the entire scene. Location could also be used to adapt the models and parameters according to different countries or even cities. Lastly, it could be used to enable a fusion between a cloud service and a safe hard real-time system. As vision processing is expensive, it is commonly offloaded to the cloud especially for augmented reality applications like Google Goggles. The location could help to adaptively send only important information based on location context.

**Sensor Fusion:** Functional safety is an important aspect of an ADAS system which necessitates the usage of additional sensors to complement the cameras or even add redundancy to obtain more confident estimates. Commonly used sensors are ultrasonics, radar, infrared and LIDAR/TOF. Ultrasonic sensors are low cost and robust for near field sensing. They are commonly employed in automated parking systems for avoiding near field collisions. Radar is suitable for mid and long range sensing and has been used commonly in ACC systems. Infrared is useful in low light scenarios to detect objects in the dark based on heat maps. LIDAR and TOF are depth sensors for estimating 3D information useful for detecting obstacles.

### B. Towards a generic scalable algorithm framework

**Motivation:** ADAS systems have grown from simple systems like LDW/PD to more sophisticated ones like TJA/HDA. This has led to a discrete component style algorithm development which are put together to form low/medium/high end systems. From a commercial perspective, this has to led to independent progression of individual modules with little re-use and scalability. As we are moving towards autonomous driving, it is beneficial to change the perspective to a top-down modelling approach. It is also useful to have this perspective for the current generation high end systems to enable joint algorithm design and re-use. Here are a few concrete examples in the different stages of the algorithm pipeline to motivate this approach. Kalman filter is commonly used for smoothing observations of objects like lanes, pedestrians, vehicles, etc. Typically they are designed and tuned individually for each object. The proposed framework encourages joint modelling where-in the common dynamics and the interactions can be captured better. Another example is the use of features for establishing correspondences for motion estimation. There are commonly used methods like Lucas-Kanade or SIFT. CNNs (deep learning) have recently performed well for these problems [29]. They are computationally very expensive and not preferred but when viewed from the perspective of a large scale system where CNNs are used for recognition of multiple objects, it becomes feasible.

**Inputs:** The inputs in our framework correspond to the physical attributes of a scene. We take the view that it could be either a raw sensor measurement or estimated from other raw values. This is inspired by how human beings perceive visual semantics. Colour information is a raw sensor measurement via rods and cones whereas motion and depth are computed in the brain. It contrasts with the classical view where only the images are the inputs and depth/motion estimation are part of the algorithm pipeline. Colour is the most basic input as it is accurately measured by a camera sensor. There are several colour formats like RGB, YUV, Lab and HSV. YUV (luma/chroma) is the common output format of cameras and for many algorithms, only the Y component is utilised. HSV is commonly used in image analysis because of the de-correlated nature of the components. Depth is another natural cue which measures distance from the camera plane. Accurate depth measurements could lead to good object segmentation. Traditionally multi-view methods like SFM (from single camera) and stereo cameras were used, but the trend is to use more accurate and expensive 3D sensors like Lidar or TOF. Depth measurements when aligned with the colour lattice gives rise to sparsity in case of both SFM (low-texture areas) and LIDAR (limited resolution and blind-spots). Hence temporal aggregation and interpolation techniques are used commonly. Motion is also a useful cue to segment out objects based on their dynamics. In the case of static objects, motion and depth are directly related; SFM uses motion estimation for computing depth. Motion is estimated by methods like Optical Flow or feature correspondence using SIFT/SURF. Background subtraction combined with odometry could also be used for segmenting the moving objects. The set of inputs finally is a collection of matrices ideally of same resolution but could be different in practise and may contain sparsity.

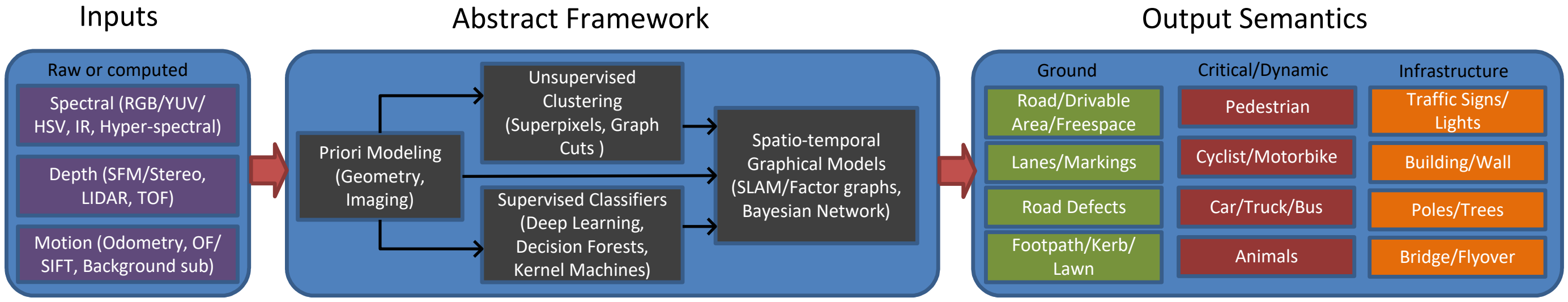

Fig. 3. Generic scalable model for vision ADAS

**Outputs:** The most generic form of output semantics is a simple matrix whose value corresponds to an object label. Sometimes it is useful to have a hierarchical labelling where-in a pixel can be labelled as Static Object → Ground → Road → Road Defect. From the perspective of automotive scenes, we have classified the outputs into three categories – ground plane objects, infrastructure and dynamic objects. Ground plane objects have a strong depth cue and could be easily detected using accurate depth measurements. Dynamic objects are the most critical, as they are movable and pose a higher risk for the driving system. They are usually learnt by supervised learning methods. All the remaining static objects are classed into infrastructure. Some of these objects need to be classified (like Traffic Signs), but most of them need to be only detected and not classified. The categories also have a relationship with the criticality of the risk involved and they are colour coded in Fig. 1 accordingly. For example, the ground surface objects pose the least risk and it is colour coded green.

**Scalable Abstract Framework:** The standard parts of the proposed framework are typical supervised and unsupervised classification methods. Supervised classifiers are commonly used for detecting objects like pedestrians, vehicles, traffic signs, etc. In this algorithm, a pattern recognition model is learnt by providing a dataset of positive and negative samples. This is scalable for problems like traffic sign recognition where there are thousands of objects wherein manual feature design becomes tedious. The classical object detection algorithms consists of a feature extraction step (HOG/LBP/Wavelets) and a kernel machine (SVM). Most of the research has been focussed on designing new features and using standard classifiers. Recently there has been a shift in this trend via deep learning methods wherein the features are automatically learned. Random forest is another popular classifier which was successful in Microsoft Kinect for pose recognition [30]. Unsupervised clustering methods group pixels based on their attributes. In case of automotive scenes, there are large blob-like structures which can be grouped together. Depth greatly simplifies the clustering algorithms and hence it is important to jointly cluster colour and depth attributes together. The popular image clustering methods are graph cuts and super-pixel segmentation. Greedy 1D traversal and merge methods are computationally efficient and effective for automotive scenes.

Context specific criteria like geometry of road, imaging models, etc., can be modelled as Bayesian priors [31]. For example, lane detection priors are its position on the ground plane, linear structure and colour contrast compared to the road. The spatio-temporal graphical model is the key part of the framework which models the spatial relationships across objects and also handles temporal associations and dynamics like birth/death of semantic segments [32]. Kerb detection is a good example where a spatial graphical model plays an important role because kerb by itself is not well defined without its spatial relationships with road, lawn, etc. SLAM is an example where the estimates are spatially and temporally refined in a statistical framework. Although the blocks in the framework are shown independent, there could be a cyclic dependency to reduce the computational complexity. The unsupervised clustering output can be used to reduce ROI for supervised classifier and temporal consistency could be exploited to use the state of spatio-temporal model to guide and refine the other models.

**Direct Trajectory Optimization:** In the field of optimal control, robotic motion is typically modelled as a trajectory optimization problem. Autonomous driving can be viewed purely as a short-term trajectory optimization based on locally observed cues [34]. Intuitively, the driver of a car does not classify all the objects and subsequently make a decision on how to manoeuvre the car; instead it is based on approximate reasoning of the environment and other motion cues, from which a global decision is taken. From a supervised classification problem perspective, the annotation is provided by recording the driver's control of the car. In terms of annotation, this method is very scalable as it can be tuned to a specific driver and a specific location. In this perspective, the problem is a straightforward supervised learning problem where the output is the estimate of the amount of steering and braking needed. There have been few successful demonstrations of this approach like Project Dave [33] and Deep driving [35]. The model learned by this approach is very abstract and could be difficult to interpret.

## VI. Conclusion

In this paper we have described for the ITS community a taxonomy of the field of computer vision in ADAS applications. We discussed some of the current and future usages of computer vision in the commercial development of autonomous vehicles. A scalable architecture for vision application development was also proposed.

Vision-based ADAS is a broad, exciting area with many challenges. In turn, this is pushing computer vision theory development. Computer vision is a field of research where it is easy to get a prototype working, but remains difficult to get an accurate system that can be used in a production vehicle. Five years ago autonomous driving seemed impossible, but after Google demonstrated their driverless car, there has been rapid progress and interest in developing the technology. Most major vehicle manufacturers are promoting some form of vehicle autonomy. Vision systems play an important role both as a detection sensor and as part of the HMI for autonomous vehicles making cameras an attractive sensor solution. Increased demands on automotive computer vision, coupled with increased image resolution (1 MPixel is current standard), the compute power of processors has increased greatly as well. Recent devices like EyeQ4 and Tegra X1 have >10x more processing power than devices used previously, such as the SOCs used in the Audi zFAS project [36]. Even if a fully autonomous mass-produced car is not achievable in the immediate future, the drive towards that goal is pushing the robustness, accuracy and performance of vision applications. To match this, government regulating bodies like EuroNCAP and NHTSA are introducing progressive legislation towards mandating safety systems, in spite of challenges in liability, and are starting to legislate to allow autonomous vehicles on the public road network.

Due to page limitations, the contents of this survey are kept high level. The authors are working on a more detailed survey that will also include hardware and software aspects of development of vision based ADAS. Algorithms could be tightly coupled with the type of hardware involved and hence the reason for the field Embedded Vision. We also cover software development processes for automotive covering functional safety, etc. The scalable statistical framework will be dealt in more detail with more formalized statistical modelling and analysis covering concrete examples.